# APPLICATIONS OF ARTIFICIAL INTELLIGENCE TECHNIQUES TO COMBATING CYBER CRIMES: A REVIEW

Selma Dilek[1], Hüseyin Çakır[2] and Mustafa Aydın[3]

[1]Department of Computer Engineering, Graduate School of Natural and Applied Sciences, Gazi University, Ankara, Turkey
[2]Department of Computer and Educational Technologies, Faculty of Education, Gazi University, Ankara, Turkey
[3]Cyber Defence and Security Center, Informatics Institute, Middle East Technical University, Ankara, Turkey

## ABSTRACT

*With the advances in information technology (IT) criminals are using cyberspace to commit numerous cyber crimes. Cyber infrastructures are highly vulnerable to intrusions and other threats. Physical devices and human intervention are not sufficient for monitoring and protection of these infrastructures; hence, there is a need for more sophisticated cyber defense systems that need to be flexible, adaptable and robust, and able to detect a wide variety of threats and make intelligent real-time decisions. Numerous bio-inspired computing methods of Artificial Intelligence have been increasingly playing an important role in cyber crime detection and prevention. The purpose of this study is to present advances made so far in the field of applying AI techniques for combating cyber crimes, to demonstrate how these techniques can be an effective tool for detection and prevention of cyber attacks, as well as to give the scope for future work.*

## KEYWORDS

*Cyber Crime, Artificial Intelligence, Intelligent Cyber Defense Methods, Intrusion Detection and Prevention Systems, Computational Intelligence*

## 1. INTRODUCTION

With the advances in information technology (IT) criminals are using cyberspace to commit numerous cyber crimes. Growing trends of complex distributed and Internet computing raise important questions about information security and privacy. Cyber infrastructures are highly vulnerable to intrusions and other threats. Physical devices such as sensors and detectors are not sufficient for monitoring and protection of these infrastructures; hence, there is a need for more sophisticated IT that can model normal behaviors and detect abnormal ones. These cyber defense systems need to be flexible, adaptable and robust, and able to detect a wide variety of threats and make intelligent real-time decisions [1, 2].

With the pace and amount of cyber attacks, human intervention is simply not sufficient for timely attack analysis and appropriate response. The fact is that the most network-centric cyber attacks are carried out by intelligent agents such as computer worms and viruses; hence, combating them with intelligent semi-autonomous agents that can detect, evaluate, and respond to cyber attacks has become a requirement. These so called computer-generated forces will have to be able to manage the entire process of attack response in a timely manner, i.e. to conclude what type of attack is occurring, what the targets are and what is the appropriate response, as well as how to prioritize and prevent secondary attacks [3].

Furthermore, cyber intrusions are not localized. They are a global menace that poses threat to any computer system in the world at a growing rate. There were times when only educated specialist could commit cyber crimes, but today with the expansion of the Internet, almost anyone has access to the knowledge and tools for committing these crimes. Conventional fixed algorithms (hard-wired logic on decision making level) have become ineffective against combating dynamically evolving cyber attacks. This is why we need innovative approaches such as applying methods of Artificial Intelligence (AI) that provide flexibility and learning capability to software which will assist humans in fighting cyber crimes [4, 5]

AI offers this and various other possibilities. Numerous nature-inspired computing methods of AI (such as Computational Intelligence, Neural Networks, Intelligent Agents, Artificial Immune Systems, Machine Learning, Data Mining, Pattern Recognition, Fuzzy Logic, Heuristics, etc.) have been increasingly playing an important role in cyber crime detection and prevention. AI enables us to design autonomic computing solutions capable of adapting to their context of use, using the methods of self-management, self-tuning, self-configuration, self-diagnosis, and self-healing. When it comes to the future of information security, AI techniques seem very promising area of research that focuses on improving the security measures for cyber space [2, 6, 7].

The purpose of this study is to present advances made so far in the field of applying AI techniques for combating cyber crimes, to demonstrate how these techniques can be an effective tool for detection and prevention of cyber attacks, as well as to give the scope for future work.

## 2. CYBER CRIMES: DEFINITION, ISSUES

The rapid development of computing technology and internet had a lot of positive impact and brought many conveniences in our lives. However, it also caused issues that are difficult to manage such as emergence of new types of crimes. For instance, common crimes such as theft and fraud attained new form of "Cyber Crimes" through information technology. Moreover, as this technology continues to evolve, criminal cases change correspondingly. Every day we are faced with increasing number and variety of cyber crimes, since this technology presents an easy way for criminals to achieve their goals. Furthermore, information technology facilitates globalization of these crimes by erasing country borders and making it much harder to monitor, detect, prevent or capture cyber criminals [8, 9, 10].

Information technology is increasingly being both targeted and used as a tool for committing crimes. Electronic devices and other high-tech products enable criminals to commit cheap and easy crimes. Computers, phones, Internet and all other information systems developed for the benefit of humanity are susceptible to criminal activity. Crimes that target information technology systems typically target e-mail accounts, bank accounts, computers, servers, websites, personal data, and digital records of private and public institutions. These crimes are also known as "Digital Crimes", "Computer Crimes", "Crimes of Information Technologies", "Network Crimes" or "Internet Crimes". Cyber crimes consist of offenses such as computer intrusions, misuse of intellectual property rights, economic espionage, online extortion, international money laundering, non-delivery of goods or services and a growing list of other offenses facilitated by Internet [8, 10, 11].

Although "cyber crime" has become a common phrase today, it is difficult to define it precisely. Most of the existing definitions were developed experimentally. Gordon and Ford (2006) define cyber crime as: "any crime that is facilitated or committed using a computer, network, or hardware device" where "computer or device may be the agent of the crime, the facilitator of the crime, or the target of the crime" [12]. Dictionary.com defines cyber crime as "criminal activity or a crime that involves the Internet, a computer system, or computer technology" [13]. Fisher and Lab (2010) defined cyber crime as "crime that occurs when computers or computer networks are involved as tool, locations, or targets of crime" [14].

Every day the amount of digital data stored and processed on computers and other computing systems increases exponentially, with people communicating, sharing, working, shopping, and socializing using computers and Internet. Language and country barriers have disappeared and virtual world has become more populated than ever. The concept of crime is present when dealing with people, therefore cyber space has not stayed isolated from the concepts of crime and criminals either [11]. Brenner (2010) argues that "most of the cyber crime we see today simply represents the migration of real-world crime to cyberspace which becomes the tool criminals use to commit old crimes in new ways." [15].

## 3. ARTIFICIAL INTELLIGENCE AND INTRUSION DETECTION

AI (also called machine intelligence in the beginning) emerged as a research discipline at the Summer Research Project of Dartmouth College in July 1956. AI can be described in two ways: (i) as a science that aims to discover the essence of intelligence and develop intelligent machines; or (ii) as a science of finding methods for solving complex problems that cannot be solved without applying some intelligence (e.g. making right decisions based on large amounts of data). In the application of AI to cyber defense, we are more interested in the second definition. Research interest in AI include ways to make machines (computers) simulate intelligent human behavior such as thinking, learning, reasoning, planning, etc. [5, 7, 16].

The general problem of simulating intelligence has been simplified to specific sub-problems which have certain characteristics or capabilities that an intelligent system should exhibit. The following characteristics have received the most attention [17, 18, 19]:

a) Deduction, reasoning, problem solving (embodied agents, neural networks, statistical approaches to AI);
b) Knowledge representation (ontologies);
c) Planning (multi-agent planning and cooperation);
d) Learning (machine learning);
e) Natural Language Processing (information retrieval – text mining, machine translation);
f) Motion and Manipulation (navigation, localization, mapping, motion planning);
g) Perception (speech recognition, facial, recognition, object recognition);
h) Social Intelligence (empathy simulation);
i) Creativity (artificial intuition, artificial imagination); and
j) General Intelligence (Strong AI).

Classic AI approaches focus on individual human behavior, knowledge representation and inference methods. Distributed Artificial Intelligence (DAI), on the other hand, focuses on social behavior, i.e. cooperation, interaction and knowledge-sharing among different units (agents). The process of finding a solution in distributed resolution problems relies on sharing knowledge about the problem and cooperation among agents. It was from these concepts that the idea of intelligent multi-agent technology emerged. An agent is an autonomous cognitive entity which understands its environment, i.e. it can work by itself and it has an internal decision-making system that acts globally around other agents. In multi-agent systems, a group of mobile autonomous agents cooperate in a coordinated and intelligent manner in order to solve a specific problem or classes of problems. They are somewhat capable of comprehending their environment, making decisions and communicating with other agents [4]. Multi-agent technology has many applications, but this study will only discuss applications to defense against cyber intrusions (See Section 4.2).

Intelligent agents systems are just a part of a much larger AI approach called Computational Intelligence (CI). CI includes several other nature-inspired techniques such as neural networks, fuzzy logic, evolutionary computation, swarm intelligence, machine learning and artificial immune systems. These techniques provide flexible decision making mechanisms for dynamic environments such as cyber-security applications. When we say 'nature-inspired', it means that there is a growing interest in the field of computing technologies to mimic biological systems (such as biological immune system) and their remarkable abilities to learn, memorize, recognize, classify and process information. Artificial immune systems (AISs) are an example of such technology [2].

AISs are computational models inspired by biological immune systems which are adaptable to changing environments and capable of continuous and dynamical learning. Immune systems are responsible for detection and dealing with intruders in living organisms. AISs are designed to mimic natural immune systems in applications for computer security in general, and intrusion detection systems (IDSs) in particular [20].

Genetic algorithms are yet another example of an AI technique, i.e. machine learning approach founded on the theory of evolutionary computation, which imitate the process of natural selection. They provide robust, adaptive, and optimal solutions even for complex computing problems. They can be used for generating rules for classification of security attacks and making specific rules for different security attacks in IDSs [21, 22].

Many methods for securing data over networks and the Internet have been developed (e.g. anti-virus software, firewall, encryption, secure protocols, etc.); however, adversaries can always find new ways to attack network systems. An intrusion detection and prevention system (IDPS) (See Fig. 1) is software or a hardware device placed inside the network, which can detect possible intrusions and also attempt to prevent them. IDPSs provide four vital security functions: monitoring, detecting, analyzing, and responding to unauthorized activities [23, 24].

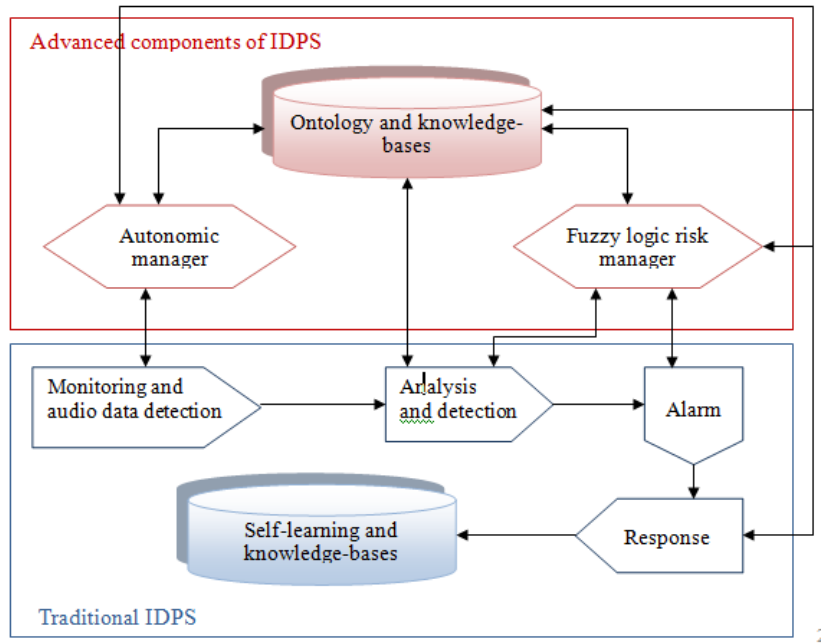

Figure 1. A typical IDPS [24].

Artificial Neural Networks (ANNs) consist of artificial neurons that can learn and solve problems when combined together. Neural networks that have ability to learn, process distributed information, self-organize and adapt, are applicable to solving problems that require considering conditionality, imprecision and ambiguity at the same time. When neural networks consist of a large number of artificial neurons, they can provide a functionality of massively parallel learning and decision-making with high speed, which makes them suitable for learning pattern recognition, classification, and selection of responses to attacks [5, 7].

## 3.1 Desired Characteristics of an IDPS

An IDPS should have certain characteristic in order to be able to provide efficient security against serious attacks. Those characteristics include the following [25]:
- Real-time intrusion detection – while the attack is in progress or immediately afterwards,
- False positive alarms must be minimized,
- Human supervision should be reduced to minimum, and continuous operation should be ensured,
- Recoverability from system crashes, either accidental or those resulting from attacks,
- Self-monitoring ability in order to detect attackers' attempts to change the system,
- Compliance to the security policies of the system that is being monitored, and
- Adaptability to system changes and user behavior over time.
-

# 4. APPLICATIONS OF AI TO DEFENSE AGAINST CYBER CRIMES

Available academic resources show that AI techniques already have numerous applications in combating cyber crimes. For instance, neural networks are being applied to intrusion detection and prevention, but there are also proposals for using neural networks in "Denial of Service (DoS) detection, computer worm detection, spam detection, zombie detection, malware classification and forensic investigations" [5]. AI techniques such as Heuristics, Data Mining, Neural Networks, and AISs, have also been applied to new-generation anti-virus technology [7]. Some IDSs use intelligent agent technology which is sometimes even combined with mobile agent technology. Mobile intelligent agents can travel among collection points to uncover suspicious cyber activity [2]. Wang et al. (2008) stated that the future of anti-virus detection technology is in application of Heuristic Technology which means "the knowledge and skills that use some methods to determine and intelligently analyze codes to detect the unknown virus by some rules while scanning" [7]. This section will briefly present related work and some existing applications of AI techniques to cyber defense.

## 4.1. Artificial Neural Network Applications

ANN is a computational mechanism that simulates structural and functional aspects of neural networks existing in biological nervous systems. They are ideal for situations that require prediction, classification or control in dynamic and complex computer environments [26].

Chen (2008) designed NeuroNet – a neural network system which collects and processes distributed information, coordinates the activities of core network devices, looks for irregularities, makes alerts and initiates countermeasures. Experiments showed that NeuroNet is effective against low-rate TCP-targeted distributed DoS attacks [27].

Linda et al. (2009) presented the Intrusion Detection System using Neural Network based Modeling (IDS-NNM) which proved to be capable of detecting all intrusion attempts in the network communication without giving any false alerts [28].

Barika et al. (2009) presented a detailed architecture of a distributed IDS based on artificial neural network for enhanced intrusion detection in networks [29].

Itikhar et al. (2009) applied neural network approach to analyzing DoS attacks. Their experiments showed that their neural network approach detects DoS attacks with more accuracy and precision than other approaches [30].

Wu (2009) presented a hybrid method of rule-based processing and back-propagation neural networks for spam filtering. Their approach proved to be much more robust compared to other spam detection approaches that use keywords, because spamming behaviors frequently change [31].

Salvador et al. (2009) presented a novel approach for Zombie PCs detection based on neural networks. Experiments showed how their approach is computationally efficient, easy to deploy in real network scenarios and achieves good Zombie detection results [32].

Bitter et al. (2010) presented host-based and network-based intrusion detection systems with a special focus on systems that employ artificial neural networks to detect suspicious and potentially malicious traffic [26].

Al-Janabi and Saeed (2011) designed a neural network-based IDS that can promptly detect and classify various attacks [33].

Barman and Khataniar (2012) also studied the development of IDSs based on neural network systems. Their experiments showed that the system they proposed has intrusion detection rates similar to other available IDSs, however, it proved to be at least 20.5 times faster in detection of DoS attacks [34].

## 4.2. Intelligent Agent Applications

Intelligent agents are autonomous computer-generated forces that communicate with each other to share data and cooperate with each other in order to plan and implement appropriate responses in case of unexpected events. Their mobility and adaptability in the environments they are deployed in, as well as their collaborative nature, makes intelligent agent technology suitable for combating cyber attacks.

Rowe (2003) developed a "counterplan" system which can prevent particular cyber attack plans using multi-agent planning together with some novel inference methods [35].

Gou et al. (2006) designed MWDCM - a multi-agent system for computer worm detection and containment in metropolitan area networks, which automatically contains the propagation of worms that waste a lot of network bandwidth and cause router crashes. The experiments showed that their system effectively thwarts worm propagation even at the high worm infection rates [36]. Phillips et al. (2006) presented a distributed agent coalition system that preserves normal operation, implements operational and security strategy, handles unforeseen events, and protects against malicious insiders, errors and attacks in distributed electric power grids [37].

Helano and Nogueira (2006) introduced a synthesis based mobile intelligent multi-agent system approach for combating cyber intrusions. They implemented their system in Prolog and applied it to combating DoS and distributed DoS attacks automatically and without human intervention [4]. Kotenko and Ulanov (2007) proposed a framework for adaptive and cooperative defense mechanisms against Internet attacks. Their approach is based on intelligent multi-agent modeling and simulation, where groups of intelligent agents interact and adjust their configuration and behavior according to the network condition and severity of attacks. They tested their approach on investigating distributed DoS attacks and defense mechanisms. The results showed that cooperation and ability to adapt in intelligent agent groups considerably raises defense effectiveness [38].

Herrero et al. (2007) presented a flexible and adaptable Mobile Visualization Connectionist agent-based IDS which system facilitates the intrusion detection in dynamic networks. In their approach, intelligent agents use artificial neural network to detect intrusions in a network [39]. Fu et al. (2007) presented an abstract model for anomaly detection in networks, inspired by biological immune system, which is based on multi-agent technology. They applied it to host and network layer in order to respond to intrusions and mitigate the damage and infection [40].

Edwards et al. (2007) researched the potential of intelligent agent technology for improving the operation and response of power grids, preventing known attacks and reducing or eliminating their consequences. They presented a MLSM – a Multi-Layered Security Model prototype that provides protection from invalid input and ability to detect and recover from unknown attack strategies (e.g. malicious input from the Internet or local disruption to the agents) [41].

Kotenko et al. (2010) studied multi-agent based approaches to the investigation and defense against botnets that are rapidly spreading across the Internet and being used to commit various cyber crimes such as performing vulnerability scans, distributed DoS attacks, and sending huge amounts of spam e-mails. They described the framework and implementation details of such systems [42].

Ye and Li (2010) presented an ad hoc network security architecture based on AIS using mobile intelligent multi-agents of two types: detection agents and counterattack agents. Their architecture, which improves security and protection of mobile ad hoc networks, combines the advantages of both AIS and intelligent agent technology and has traits of distribution, self-adaption, self-learning, and expandability [43].

Wei et al. (2010) proposed a theoretical layered approach for protecting power grid automation systems against cyber attacks which may come from either the Internet or internal networked sources. One of the components of their framework are security agents some of which are intelligent and have ability to detect intrusive events and activities within the controllers. The results obtained by testing a prototype of the proposed approach showed that their system is able to manage and mitigate some common vulnerability issues of power grid automation systems [44].

Doelitzscher et al. (2011) introduced SAaaS – a cloud incident detection system Security Audit as a Service. Their system is based on intelligent autonomous agents which are aware of underlying business flows of deployed cloud instances; thus providing flexibility and supported cross customer event monitoring of a cloud infrastructure [45].

Shosha et al. (2011) proposed a distributed IDS based on a community collaboration between multiple agents for detecting cyber intrusions in Supervisory Control and Data Acquisition (SCADA) networks. The proposed architecture also incorporates the SCADA network topology and connectivity constraints [46].

Ionita and Ionita (2013) proposed a multi intelligent agent based approach for network intrusion detection using data mining [47].

## 4.3. Artificial Immune System Applications

AISs, just like the biological immune systems which they are based on, are employed to uphold stability in a changing environment. The immune-based intrusion detection comprises the evolution of immunocytes (self-tolerance, clone, variation, etc.) and antigens detection simultaneously. An immune system produces antibodies to resist pathogens and the intrusion intensity can be estimated by variation of the antibody concentration. Therefore, AISs play an important role in the cyber security research [48].

Sirisanyalak and Sornil (2007) presented an AIS-based e-mail feature extraction approach for spam detection. The performance evaluation results showed that the proposed method is much more efficient in spam detection than other existing systems, with very low false positive and false negative rates (0.91% and 1.95% respectively) [49].

Lebbe et al. (2007) analyzed various AIS models used in IDSs and introduced Danger Theory (DT) in AIS as a method for danger response in wireless mesh networks. For classification of network dangers they used Self-organizing Maps (SOMs) as classifiers. Their experiments validated their proposal of applying DT to security of wireless mesh networks [50].

Hong (2008) presented an AIS-based hybrid learning algorithm for anomaly detection in computer systems [20].

Gianini et al. (2009) proposed an extension of AIS model for computer system security to ambient intelligence domain. Their extended model can provide perceptual functions and detection capabilities with device intelligence (e.g. multimedia sensor system interpretation) [51]. EshghiShargh (2009) also studied the benefits of AI in general, and AISs in particular, for improving IDSs by investigating different IDS designs based on AISs. The results showed how AIS approach to IDS design can be fruitful for future applications [52].

Chao and Tan (2009) proposed a novel virus detection system based on AIS. The experimental results showed that virus detection system they proposed has a "strong detection ability and good generalization performance" [53].

Danforth (2009) investigated the possibility of expanding AISs for classification of web server attacks, which could help system administrator with a warning about the severity of the attack and assist in mitigation of direct attacks [54].

Mohamed and Abdullah (2009) presented an AIS-based security framework for securing mobile ad hoc networks, which is scalable, robust, and has traits of distributability, second response and self-recovery. Their architecture resolved some limitations found in the previous related studies such as scalability and bandwidth conservation [55].

Qiang and Yiqian (2010) proposed an AIS-based network security situation assessment model which can make real-time and quantitative security situation assessment of the system, and provide support required to make real-time adjustments of the defense measures. Theoretical analysis and experiments showed the effectiveness of the model in real-time anomaly detection for network security [56].

Rui and Wanbo (2010) proposed an AIS-based self-learning intrusion response model which can recognize and classify unknown attacks. Their model has a dynamic response decision-making mechanism which can adjust the defensive tactics according to the changes in the environment and keep the system safe with the minimum cost. The experiments showed that their model has qualities such as self-adaptation, rationality, quantitative calculation, and that it provides efficient intrusion response [48].

Endy et al. (2010) used SOMs to visualize the topology of the data in order to perform cluster analysis of the textual documents related to cyber terrorism [57].

Yang et al. (2011) presented a network security evaluation model for quantitative analysis of the degree of intrusion danger level based on AIS theory, and demonstrated its advantages over traditional models for network security evaluation [58].

Liu et al. (2011) introduced an AIS-based intrusion detection mechanism into the Internet of Things (IoT) environment, which simulates self-adaptation and self-learning mechanisms via dynamic adaptation to the environment. The analysis of their proposal showed that their model provides a new effective intrusion detection for the Internet of Things [59].

Zhang et al. (2011) proposed SGDIDS – a new hierarchical Distributed Intrusion Detection System for improving cyber security of the Smart Grid. Their system consists of an intelligent module (among other modules) which uses AIS to detect and classify malicious data and possible cyber attacks. Simulation results showed that their system is applicable to identification of malicious network traffic and improving system security [60].

Ansari and Inamullah (2011) proposed an enhancement for the anomaly detection based on AIS and showed how their model improves AIS performance in applications such as anomaly detection, ensuring security, detecting errors and performing data mining in mobile ad hoc networks [61].

Fang et al. (2012) proposed an AIS for phishing detection through memory and mature detectors. The analysis showed that their system is unique and more flexible and adaptive than other existing phishing detection systems [62].

Mavee and Ehlers (2012) proposed IISGP – a new AIS-based model for Smart Grid protection. Basically, they aimed to design a bio-inspired AIS model for intrusion detection, access control and anomaly detection in critical infrastructures which are becoming increasingly dependent on cyber technology [63].

Kumar and Reddy (2014) developed a unique agent based intrusion detection system for wireless networks that collects information from various nodes and uses this information with an evolutionary AIS to detect and prevent the intrusion via bypassing or delaying the transmission over the intrusive paths. The experimental results showed that the system is well suited for intrusion detection and prevention in wireless networks [64].

## 4.4. Genetic Algorithm and Fuzzy Sets Applications

Kim et al. (2004) proposed a learning algorithm for anomaly detectors which can detect attacks using genetic algorithm. They applied their algorithm to an artificial computer security system and showed its effectiveness in intrusion detection [65].

Sekeh and Bin Maarof (2009) proposed a fuzzy host-based intrusion detection system that uses data mining technique and the services of the underlying operating system calls. The simulation results showed that the proposed system improves the performance, and decreases the size of the database, time complexity, and the rate of false alarms [66].

Mabu et al. (2011) described a novel fuzzy network intrusion detection method based on class-association-rule mining in genetic network programming. The proposed method is flexible and efficient for both misuse and anomaly detection in networks and it is capable of dealing with the mixed databases which contain both discrete and continuous attributes to mine important class-association rules needed for improved intrusion detection. The experiments and evaluation of the proposed method demonstrated that this approach provides competitively high detection rates in comparison with other machine-learning techniques [67].

Ojugo et al. (2012) presented GAIDS – a Genetic Algorithm Rule-Based Intrusion Detection System for improving system security, confidentiality, integrity and resource availability in networked settings. The proposed system uses a set of classification rules obtained from network audit data and the support-confidence framework, used as fitness function to evaluate the quality of each rule [68].

Hassan (2013) designed an IDS based on genetic algorithm and fuzzy logic for efficient detection of various intrusive activities within a network. The system is adaptable and cost-effective as it can update rules once new intrusive activities become known. The experiments and evaluations results showed that the proposed system achieved reasonable intrusion detection rate [69].

Jongsuebsuk et al. (2013) proposed a network IDS based on a fuzzy genetic algorithm. Fuzzy rules are used to classify network attack data, whereas genetic algorithm optimizes finding appropriate fuzzy rule in order to obtain the optimal solution. The evaluation results showed that the proposed IDS can detect network attacks in real-time (or within 2-3 seconds) upon the arrival of data arrives to the detection system with the detection rate of over 97.5% [70].

Chaudhary et al. (2014) developed an anomaly based fuzzy intrusion detection system to detect the packet dropping attacks in mobile ad hoc networks. The simulation results demonstrated that proposed system has the capability to detect the packet dropping attacks with high positive and low false positive rates under all speed levels of mobile nodes [71].

Benaicha et al. (2014) presented a network intrusion detection model based on Genetic Algorithm approach with an improved initial population and selection operator used to optimize the search of attack scenarios in audit files and provide the subset of potential attacks within realistic processing time. They employed genetic algorithm approach because it boosts the performance and reduces the false positive rate [72].

Padmadas et al. (2014) presented a layered genetic algorithm-based intrusion detection system for monitoring activities in a given environment to determine whether they are legitimate or malicious based on the available information resources, system integrity and confidentiality. The experimental results showed that the proposed system efficiently detect R2L attacks with 90% accuracy [73].

## 4.5. Other AI Applications

In this section we give examples of various other and hybrid applications.
Machado et al. (2005) presented a novel network intrusion detection model based on mobile intelligent agent technology and AISs. They also implemented their design and showed that it is capable of differentiating between various attacks, security violations, and several other security breaches. The experimental results showed that their model offers a significant upgrade compared to previous work in the field [74].

Pei and Song (2008) focused on improving the performance of intrusion detectors of IDSs, so they proposed a hybrid approach which uses the searching performance of immune algorithm to generate fuzzy-detectors. The experiments showed the great searching capability of immune algorithm. Results also showed that fuzzy detection rules reduce the frangibility of detectors and improve the detection precision [75].

Zhou (2009) proposed a method for merging AIS technique and neural networks to construct an intrusion detection model capable of both anomaly detection and misuse detection. Evaluation and experimental results showed high intrusion detection accuracy with low false alarm rate [76]. Golovko et al. (2010) also proposed using AISs and neural networks for attack detection in computer systems. They described principles and architecture of such an attack detection system [77].

Elsadig et al. (2010) described a novel approach for bio-inspired intrusion prevention and self-healing system. They presented a novel AIS-based intrusion prevention system (IPS) which uses intelligent multi-agent system for non-linear classification method to identify the abnormality behavior and detect, prevent and heal harmful or dangerous events network system [78].

Zhou et al. (2011) presented an AIS-based IDS for combating virus with "virus". They implanted "virus" and cloned variation of "virus" into immune IDS based on e-learning in order to improve immunity of the system and eliminate invasion or attack behaviors [79].

Ou et al. (2011) proposed ABAIS - a multi-agent based AIS for IDSs with learning and memory capabilities. Immune response to malicious activity is activated by either computer host or security operating center. Experimental results showed that ABAIS can effectively detect malicious intrusions [80].

Meng (2011) researched the holistic intelligence of Neuro-Endocrine-Immune system and presented an artificial homeostasis security-coordination model. A prototype of the model was implemented for E-Government system. The study showed that artificial homeostasis model can integrate different security products to coordinate in intrusion detection, security management, and prevention of potential attacks or system security vulnerabilities [81].

Dove (2011) researched abnormal behavior detection and the limitations of looking only for known attack patterns in cyber domain and suggested that these issues can be addressed by having a model that engages in continuous learning and re-profiling of normal behavior and uses a sense-making hierarchy to reduce false positive rates. The architecture is based on process patterns inspired by biological immune system combined with hierarchical sense-making [82].
Jiang et al. (2011) proposed a bio-inspired host-based multilayered intrusion detection system using multiple detection engines and sequential pattern recognition. The results showed that their model efficiently classifies unknown behaviors and malicious attacks and that it can successfully identify the region where abnormalities are likely to occur with lower false positive rate compared to other existing schemes. They also argued that their study provides the basis for an intelligent and computationally simple real-time approach for detecting unknown malware and malicious attacks in large-scale complex networks [83].

Ferreira et al. (2011) presented an IDS based on the wavelet and ANN that is applied to the well know Knowledge Discovery and Data Mining KDD. Their experiment showed high intrusion detection rate [84].

Wattanapongsakorn et al. (2012) presented a simple network-based intrusion detection and prevention system (IDPS) which uses several machine learning algorithms to detect and classify network attacks. They tested it in an online network environment and the results showed that the proposed IDPS offers high detection rate for the main attack types (Probe and DoS) within a few seconds, and also automatically protects the computer network from the attacks. It also worked well for unknown types of network attacks [85].

Aziz et al. (2012) developed an AIS-inspired [85] network intrusion detection system is which uses detectors generated by a genetic algorithm combined with deterministic-crowding niching technique. They achieved an overall average detection rate of 81.74% [86].

Patel et al. (2013) presented a state-of-the-art IDPSs with possible solutions to intrusion detection and prevention in cloud computing which is an attractive target for potential cyber attacks. They identified relevant requirements for an ideal cloud based IDPS: autonomic computing self-management, ontology, risk management, and fuzzy theory [24].

Barani (2014) proposed GAAIS – a dynamic intrusion detection method for Mobile ad hoc Networks based on genetic algorithm and artificial AIS. GAAIS is self-adaptable to network topology changes. The performance of the proposed system was evaluated for detection of several types of routing attacks such as Flooding, Blackhole, Neighbor, Rushing, and Wormhole attacks. The experimental results demonstrated that it is more efficient when compared to similar approaches [87].

## 4.6. Advantages of AI Applications to IDPSs

AI techniques introduce numerous advantages into intrusion detection and prevention (see Table 1).

Table 1. Advantages that some AI techniques bring to intrusion detection and prevention.

| Technology | Advantages |
|---|---|
| **Artificial Neural Networks** | Parallelism in information processing;<br>Learning by example;<br>Nonlinearity – handling complex nonlinear functions;<br>Superiority over complex and perplexing differential equations;<br>Resilience to noise and incomplete data;<br>Versatility and flexibility with learning models;<br>Intuitiveness – as they are an abstraction of biological neural networks [26]. |
| **Intelligent Agents** | Mobility;<br>Helpfulness – they always attempt to accomplish their tasks having contradictory objectives;<br>Rationality – in achieving their objectives;<br>Adaptability – to the environment and user preferences;<br>Collaboration – awareness that a human user can make mistake and provide uncertain or omit important information; thus, they should not accept instructions without considerations and checking the inconsistencies with the user [4]. |
| **Artificial Immune Systems** | Dynamic structure;<br>Parallelism and distributed learning – using data network communications and parallelism in detection and elimination tasks;<br>Self-adaptability and self-organizing – updating intrusion marks without human involvement; |

| | Robustness; Selective response – removing malicious activity by the best means available; Diversity – each detector node generates a statistically unique set of non-self detectors; Resource optimization; Multi-layered structure – attackers cannot succeed with their malicious activities by circumventing only one layer, since multiple layers of different structures are in charge of monitoring a single point. Disposability – not being dependent on a single component which can be easily replaced by other components [52, 56, 88]. |
|---|---|
| **Genetic Algorithms** | Robustness; Adaptability to the environment; Optimization – providing optimal solutions even for complex computing problems; Parallelism – allowing evaluation of multiple schemas at once; Flexible and robust global search [21, 86]. |
| **Fuzzy Sets** | Robustness of their interpolative reasoning mechanism; Interoperability – human-friendliness [89, 90]. |

# 5. LIMITATIONS OF CURRENT ANOMALY DETECTION/PREVENTION SYSTEMS

Although anomaly detection systems offer the opportunity to detect previously unknown attacks, they have some important limitations that need to be tackled. The main issue is the difficulty of making a solid model of what acceptable behavior is and what an attack is; hence, they may give a high number of false positive alarms, which may be caused by atypical behavior that is actually normal and authorized, since normal behavior may easily and readily change. Other limitations include the following [25, 26, 29]:

- In order for the anomaly detection system to be able to characterize normal patterns and create a model of the normal behavior, wide-ranging training sets of the normal system activities are needed. Any change in the system's normal patterns must lead to necessary update of the knowledge base.
- If the detection and prevention system inaccurately classifies a legitimate activity as a malicious one, the results can be very unfortunate since it will attempt to stop the activity or change it.
- An intrusion detection system, no matter how efficient, may be disabled by attackers if they can learn how the system works.
- In heterogeneous environments there is also an issue of integrating information from different sites.
- Another problem involves supplying intrusion detection systems that will conform to legal regulations, security requirements and/or service-level agreements in real world.

# 6. SCOPE FOR FUTURE WORK

Cyber security needs much more attention. Given human limitations and the fact that agents such as computer viruses and worms are intelligent, network-centric environments require intelligent cyber sensor agents (or computer-generated forces) which will detect, evaluate and respond to cyber attacks in a timely manner [3].

Application of AI techniques in cyber defense will need planning and future research. One of the challenges is knowledge management in network-centric warfare, hence a promising area for research is introduction of modular and hierarchical knowledge architecture in the decision making software. Rapid situation assessment and decision superiority can only be guaranteed with automated knowledge management. It is also foreseeable that the grand goal of AI research – development of artificial general intelligence - can be reached in not so distant future which would lead to Singularity described as "the technological creation of smarter-than-human intelligence". Nevertheless, it is of crucial importance that we have the ability to use better AI technology in cyber defense than the one offenders possess [5].

Furthermore, a lot more research needs to be done before we are able to construct trustworthy, deployable intelligent agent systems that can manage distributed infrastructures. Future work must search for a theory of group utility function to allow groups of agents to make decisions [37].

For future work in enhancing IDPSs, unsupervised learning algorithms and new techniques will be considered together to create hybrid IDPS which will improve the performance of anomaly intrusion detection [85]. Moreover, combining all kinds of AI technologies will become the main development trend in the field of anti-virus technology [7].

Even though computational intelligence techniques have been widely used in the field of computer security and forensics, there are certain ethical and legal problems that arise as the technology rapidly expands. Some of these problems are privacy concerns or power issues on the ethical side or questions of due process on the legal side. A wide range of both ethical and legal questions come up in the light of the potential autonomy of this technology. Questions like "to what extent can an artificial neural network replace human judgment", "to what degree do we want to allow technology to take human roles" or "what legal precedent can be applied to machines" will need to be answered [91].

# 7. CONCLUSION

The fast development of information technology had a lot of positive impact and brought many conveniences into our lives. However, it also caused issues that are difficult to manage such as the emergence of cyber crimes. As the technology continues to evolve, criminal cases change correspondingly. Every day we are faced with increasing number and variety of cyber crimes, since this technology presents an easy way for criminals to achieve their goals. Critical infrastructures are especially vulnerable.

Application of AI techniques are already being used to assist humans in fighting cyber crimes, as they provide flexibility and learning capabilities to IDPS software. It has become obvious wide knowledge usage in decision making process requires intelligent decision support in cyber defense which can be successfully achieved using AI methods.

Available academic resources show that AI techniques already have numerous applications in combating cyber crimes. This paper has briefly presented advances made so far in the field of applying AI techniques for combating cyber crimes, their current limitations and desired characteristics, as well as given the scope for future work.

**Authors**

Selma Dilek is a graduate student in Computer Engineering at the Graduate School of Natural and Applied Sciences, Gazi University, Ankara, Turkey. She obtained her B.Sc. in Computer Science (major) and Electrical and Electronic Engineering (minor) from the University Sarajevo School of Science and Technology, Sarajevo, Bosnia and Herzegovina, and B.Sc. in Computer Science from Buckingham University, UK. Her publications have appeared in an international journal and conference proceedings. Her research interests include Wireless Sensor Networks, Network Security, Operating Systems, Algorithm Design, Robotics, and Artificial Intelligence.

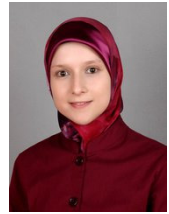

Hüseyin Çakır is an assistant professor of Computer and Educational Technologies, at the Faculty of Education, Gazi University, Ankara, Turkey. He obtained his Bachelor's and Master's degree in Computer Education from Gazi University, and PhD in Educational Technologies from Ankara University, Turkey. His publications have appeared in various national and international journals and conference proceedings. His research interests include Computer Hardware and Software, Educational Software, Computer Assisted Education, Web Based Education, Visual Programming, Internet and Web Technologies, 3D Graphics, Animation, Educational Technology, Multiple Intelligences, Traffic Education, Forensic Computing, and Artificial Intelligence.

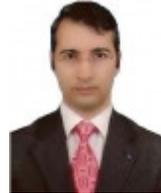

Mustafa Aydın is a PhD student in Information Systems at the Graduate School of Informatics, Middle East Technical University, Ankara, Turkey. He obtained his B.Sc. and M.Sc. degrees in Electrical and Electronics Engineering from Gazi University and Hacettepe University, Turkey, respectively. He is also a member of Cyber Defence and Security Center of Middle East Technical University. His research interests include Information Technologies and Systems Security, Information Technology Management and Audit, Cyber Defence, Machine Learning, Data Mining and Statistics.

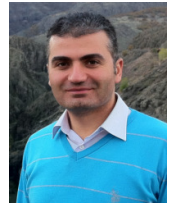